\documentclass[times, 10pt,twocolumn]{article}
\usepackage{latex8}
\usepackage{times}

\usepackage{amsmath,epsfig}
\usepackage{amsfonts}
\usepackage{balance}
\usepackage{amssymb}
\usepackage{graphicx}
\usepackage{euscript}
\usepackage{algorithm, algorithmic}
\usepackage{color}
\usepackage{breqn}
\usepackage{subfigure}
\usepackage{multirow,hhline}

\begin{document}

\title{Do GANs leave artificial fingerprints?}

\author{Francesco Marra, Diego Gragnaniello, Luisa Verdoliva, Giovanni Poggi\\
DIETI -- University Federico II of Naples\\
Via Claudio 21, 80125 Napoli -- ITALY\\
\{francesco.marra, diego.gragnaniello, verdoliv, poggi\}@unina.it\\
}

\maketitle
\thispagestyle{empty}

\begin{abstract}
In the last few years,
generative adversarial networks (GAN) have shown tremendous potential for a number of applications in computer vision and related fields.
With the current pace of progress, it is a sure bet they will soon be able to generate high-quality images and videos, virtually indistinguishable from real ones.
Unfortunately, realistic GAN-generated images pose serious threats to security, to begin with a possible flood of fake multimedia,
and multimedia forensic countermeasures are in urgent need.
In this work, we show that each GAN leaves its specific fingerprint in the images it generates,
just like real-world cameras mark acquired images with traces of their photo-response non-uniformity pattern.
Source identification experiments with several popular GANs
show such fingerprints to represent a precious asset for forensic analyses.
\end{abstract}

\section{Introduction}

Generative adversarial networks are pushing the limits of image manipulation.
A skilled individual can easily generate realistic images sampled from a desired distribution
\cite{Salimans2016, Gulrajani2017, Berthelot2017},
or convert original images to fit a new context of interest \cite{Thies2016, Isola2017, Zhu2017, Liu2017, Choi2018}.
With progressive GANs \cite{karras2018}, images of arbitrary resolution can be created, further improving the level of photorealism.

\begin{figure}[t!]
	\centering
	\begin{tabular}{c@{\hspace{1mm}}c@{\hspace{1mm}}c}
		\includegraphics[width=0.30\columnwidth]{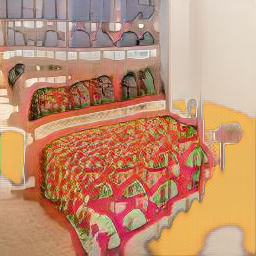} &
		\includegraphics[width=0.30\columnwidth]{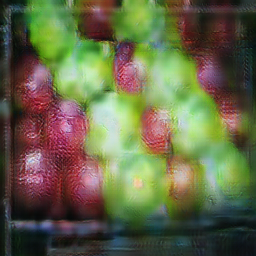} &
		\includegraphics[width=0.30\columnwidth]{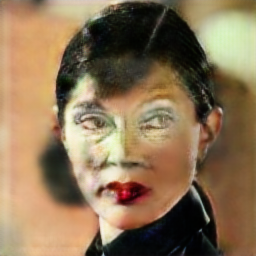} \\

		\includegraphics[width=0.30\columnwidth]{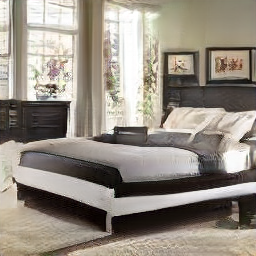} &
		\includegraphics[width=0.30\columnwidth]{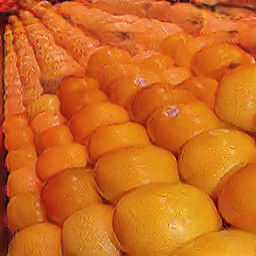} &
		\includegraphics[width=0.30\columnwidth]{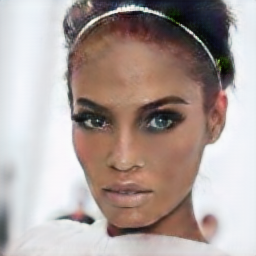} \\
    (a) & (b) & (c) \\
	\end{tabular}
	\caption{\small Sample images generated by Pro-GAN (a), Cycle-GAN (b), Star-GAN. Top: easily detected bad results. Bottom: photorealistic results.}
	\label{fig:GANExamples}	
\end{figure}

There is widespread concern on the possible impact of this technology in the wrong hands.
Well-crafted fake multimedia add further momentum to the already alarming phenomenon of fake news,
if ``seeing is believing'', as they say.
Although today's GAN-based manipulations present often artifacts that raise the suspect of observers, see Fig.1(top),
this is not always the case (bottom),
and it is only a matter of time before GAN-generated images will consistently pass visual scrutiny.
Therefore, suitable multimedia forensic tools are required to detect such fakes.

In recent years, a large number of methods have been proposed to single out fake visual data,
relying on their semantic, physical, or statistical inconsistencies \cite{Farid2016}.

\begin{figure*}
	\centering
	\begin{tabular}{c@{\hspace{1mm}}c@{\hspace{1mm}}c@{\hspace{1mm}}c@{\hspace{1mm}}c}
		\includegraphics[width=32mm,height=32mm]{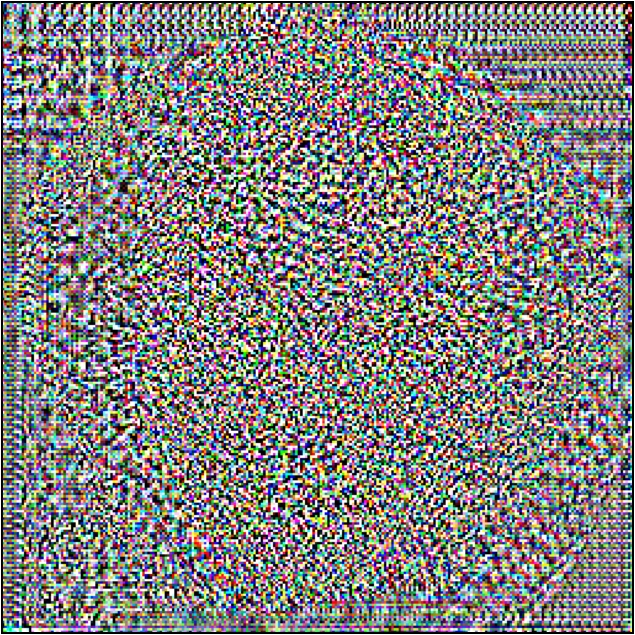} &
		\includegraphics[width=32mm,height=32mm]{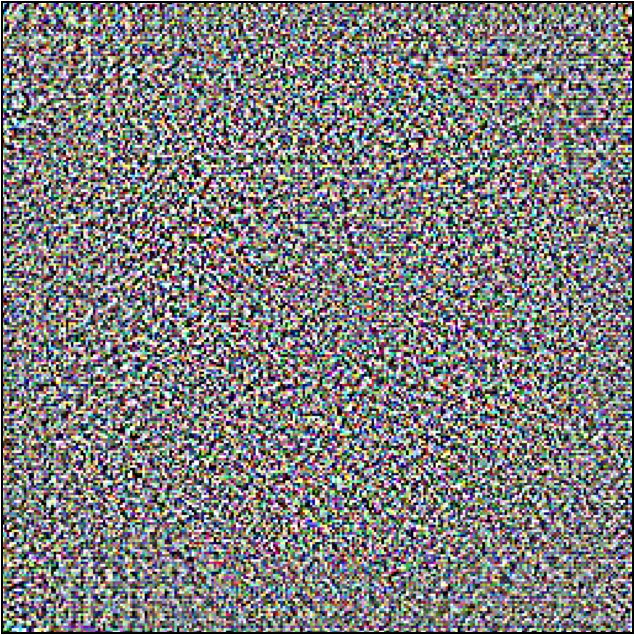} &
		\includegraphics[width=32mm,height=32mm]{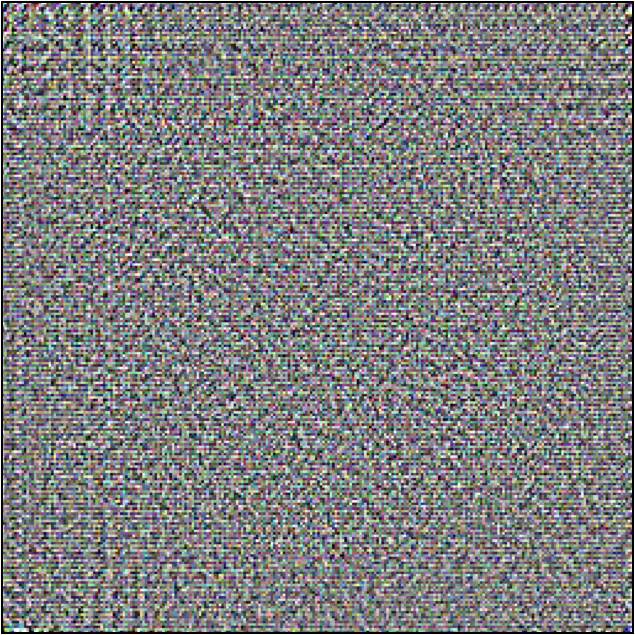} &
		\includegraphics[width=32mm,height=32mm]{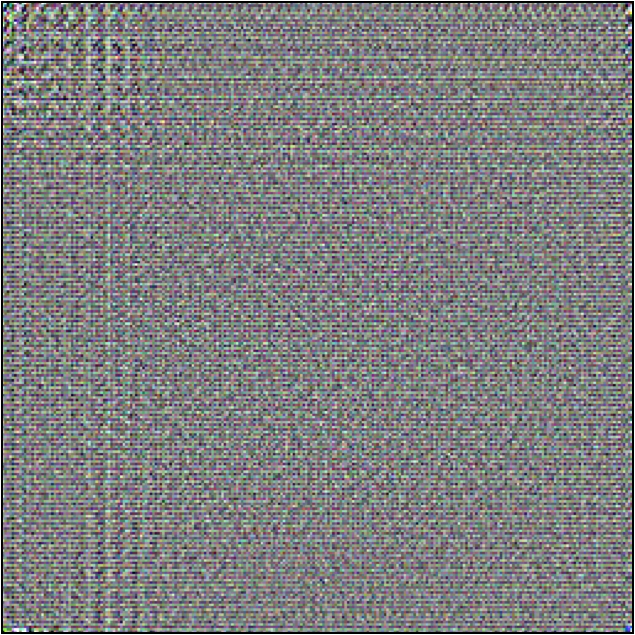} &
		\includegraphics[width=32mm,height=32mm]{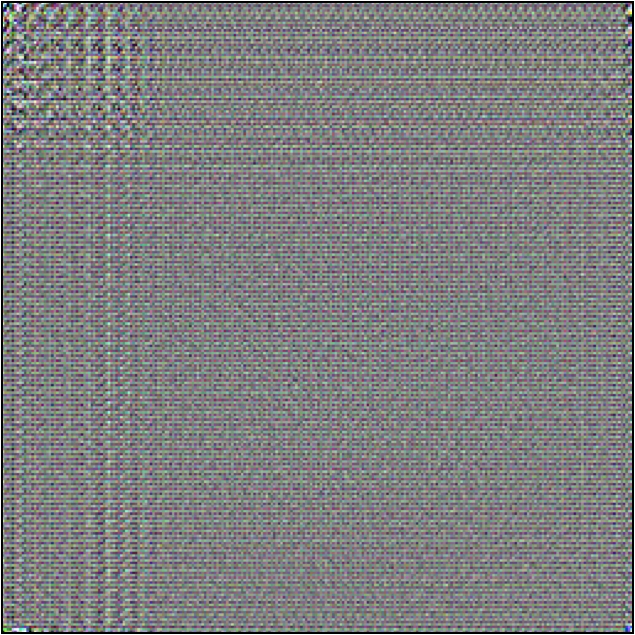} \\
		\includegraphics[width=32mm,height=32mm]{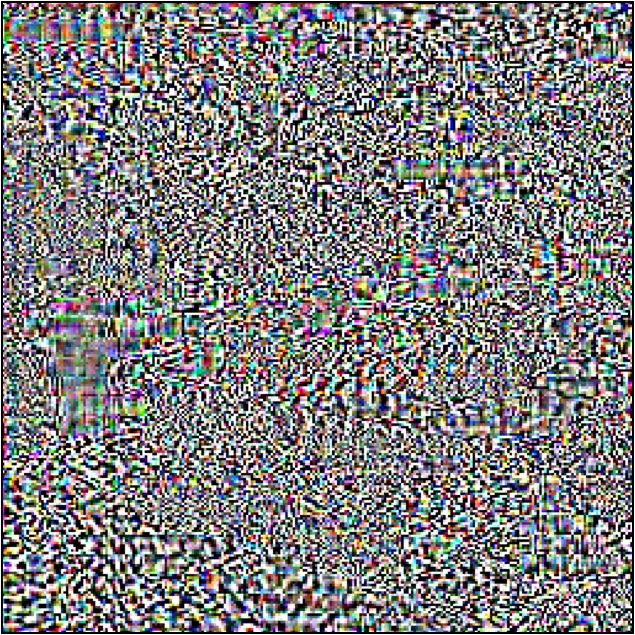} &
		\includegraphics[width=32mm,height=32mm]{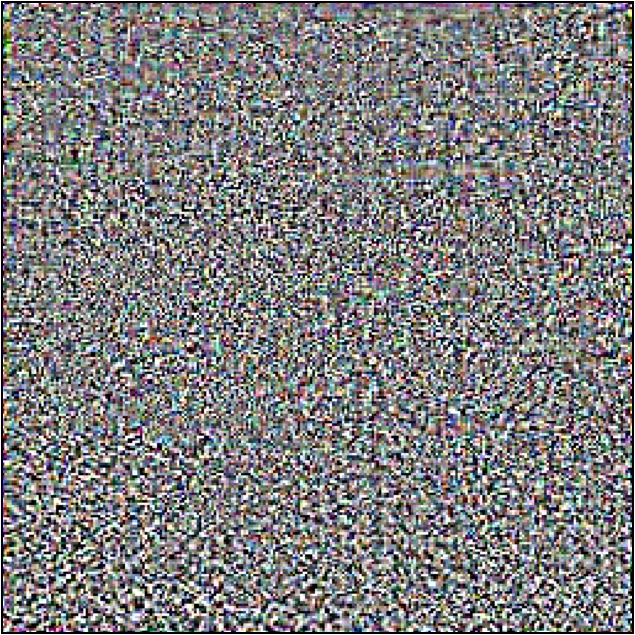} &
		\includegraphics[width=32mm,height=32mm]{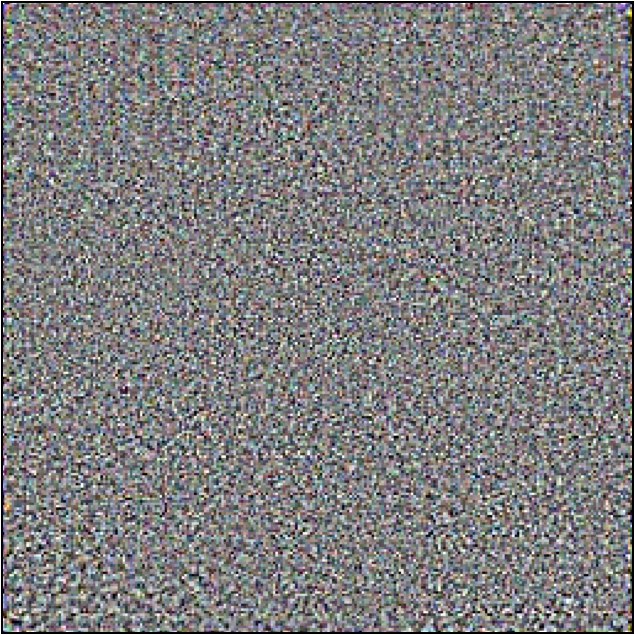} &
		\includegraphics[width=32mm,height=32mm]{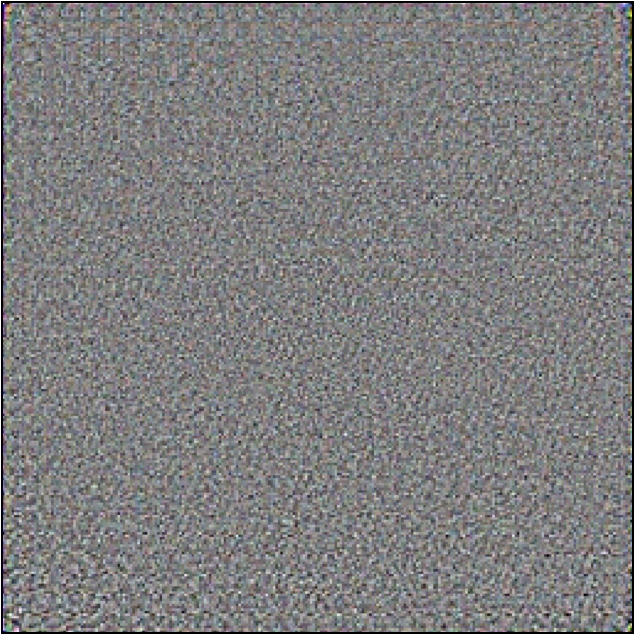} &
		\includegraphics[width=32mm,height=32mm]{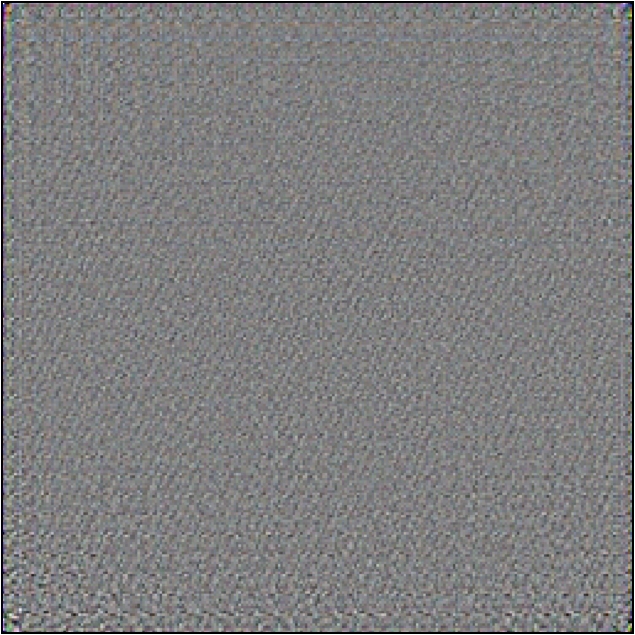} \end{tabular}
	\caption{\small Cycle-GAN o2a (top) and Pro-GAN kitchen (bottom) fingerprints estimated with 2, 8, 32, 128, 512 residuals.}
	\label{fig:GAN_fingerprints}	
\end{figure*}

Statistical-based approaches, in particular, rely on the long trail of subtle traces left in each image by the acquisition devices,
traces that can be hardly disguised even by a skilled attacker.
In fact,
each individual device, due to manufacturing imperfections, leaves a unique and stable mark on each acquired photo,
the photo-response non-uniformity (PRNU) pattern \cite{Lukas2006},
which can be estimated and used as a sort of {\em device} fingerprint.
Likewise,
each individual acquisition model, due to its peculiar in-camera processing suite (demosaicking, compression, etc.),
leaves further model-related marks on the images, which can be used to extract a {\em model} fingerprint \cite{Cozzolino2018}.
Such fingerprints can be used to perform image attribution \cite{Lukas2006, Chen2008},
as well as to detect and localize image manipulations \cite{Chen2008, Cozzolino2018},
and represent one of the strongest tools in the hands of the forensic analyst.

GANs have little in common with conventional acquisition devices,
and GAN-generated images will not show the same camera-related marks.
Still, they are the outcome of complex processing systems involving a large number of filtering processes,
which may well leave their own distinctive marks on output images.
So the question\footnote{Winking at P.K.Dick novel: ``Do androids dream of electric sheep?''} is: do GANs leave artificial fingerprints?
That is, do the images generated by a given GAN share a common and stable pattern that allows to establish their origin?
And, if this is the case, how reliable will such a fingerprint be? How robust to defensive measures? And how discriminative about the image origin?

In this paper we investigate on this interesting issue, and provide a first answer to the above questions.
Our experiments with several popular GAN architectures and datasets,
show that GAN do leave specific fingerprints on the image they generate, which can be used to carry out reliable forensic analyses.

\section{Related Work}

Recently there has been a growing interest in distinguishing GAN-generated images from real ones.
As shown in Fig.1, the current state of the art in GANs is far from perfection, and often generated images exhibit strong visual artifacts that can be exploited for forensic use.
For example, to detect fake faces, \cite{Matern2019} exploits visual features regarding eyes, teeth and facial contours.
Tellingly, the authors observe that in GAN-generated images the color of left and right eye are often inconsistent.
Color information is also used in \cite{McCloskey2018, Li2018}.
In particular, \cite{McCloskey2018} proposes to use some features shared by different GAN architectures, related to the synthesis of RGB color channels.
Other methods rely on deep learning.
Several architectures have been tested so far \cite{Marra2018, Mo2018, Tariq2018} showing a good accuracy in detecting GAN-generated images, even on compressed images.
Unfortunately, if a network is trained on a specific architecture,
its performance degrades sharply when used to detect image generated by another architecture \cite{Cozzolino2018FT}.
This observation suggests the presence of different artifacts peculiar of each specific GAN model.
Recently, it has also been shown \cite{Yu2018} that a deep network can reliably discriminate images generated with different architectures.
However, the network requires intensive training on an aligned dataset,
and there is no hint, let alone exploitation, of the presence of GAN-induced fingerprints.

\section{Exposing GAN fingerprints}

In this Section we show evidence on the existence of GAN fingerprints.
This goal is pursued in a minimal experimental setting,
considering only two GANs, a Cycle-GAN trained to convert orange images into apple images and a Progressive-GAN (Pro-GAN) trained to generate kitchen images,
call them GAN A and B, from now on.
Lacking any statistical model, we consider an extraction pipeline similar to that of the PRNU pattern.
For the generic image $X_i$ generated by a given GAN, the fingerprint represents a disturbance, unrelated with the image semantics.
Therefore,
we first estimate the high-level image content, $\widehat{X}_i=f(X_i)$, through a suitable denoising filter $f(\cdot)$,
then subtract it from the original image to extract the noise residual
\begin{equation}
    R_i = X_i-f(X_i)
\end{equation}
Then, we assume the residual to be the sum of a {\em non-zero} deterministic component, the fingerprint $F$, and a random noise component $W_i$
\begin{equation}
    R_i = F+W_i
\end{equation}
Accordingly, the fingerprint is estimated by a simple average over the available residuals
\begin{equation}
    \widehat{F} = \frac{1}{N} \sum_{i=1}^N R_i
\end{equation}
Fig.2 shows (suitably amplified) the fingerprints of the two GANs, estimated over a growing number of residuals, $N=2,8,32,128,512$.
Of course, for low values on $N$, the estimates are dominated by image-related noise.
However, as $N$ grows, the additive noise component tends to vanish and both estimates converge to stable quasi-periodical patterns,
which we regard as accurate approximations of the true GAN fingerprints.
In Fig.3 we show the energy $E(N)$ of these estimated fingerprints as a function of $N$,
together with the best fitting curve of the type
\begin{equation}
    \widehat{E}(N) = E_{\infty}+E_0\times2^{-N}
\end{equation}
The fitting is clearly very accurate for large values of $N$, and the $E_{\infty}$ value estimates the energy of the limit fingerprint, 0.0377 and 0.0088, respectively.
Fig.4, instead, shows the autocorrelation functions of the two estimates for $N$=512,
with clear quasi-periodical patterns providing further evidence of the non-random nature of these signals.

\begin{figure}
	\centering
	\begin{tabular}{c@{\hspace{2mm}}c}
		\includegraphics[width=40mm,height=36mm]{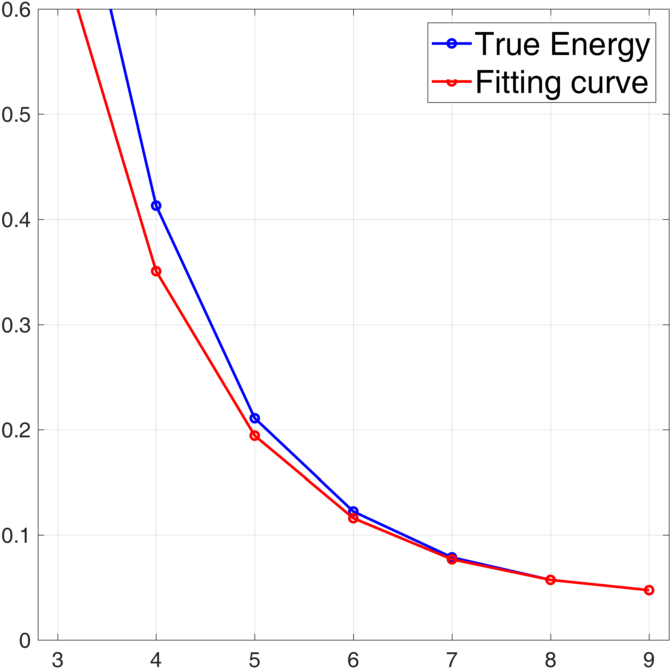} &
		\includegraphics[width=40mm,height=36mm]{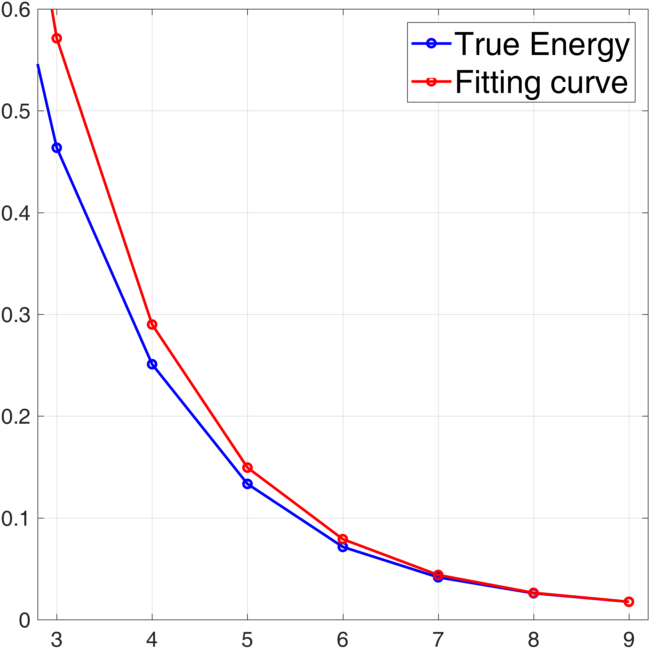}
    \end{tabular}
	\caption{\small True energy and fitting curve for the Cycle-GAN and Pro-GAN fingerprints of Figure 2.}
	\label{fig:Energies}	
\end{figure}

\begin{figure}
	\centering
	\begin{tabular}{c@{\hspace{2mm}}c}
		\includegraphics[width=40mm,height=36mm]{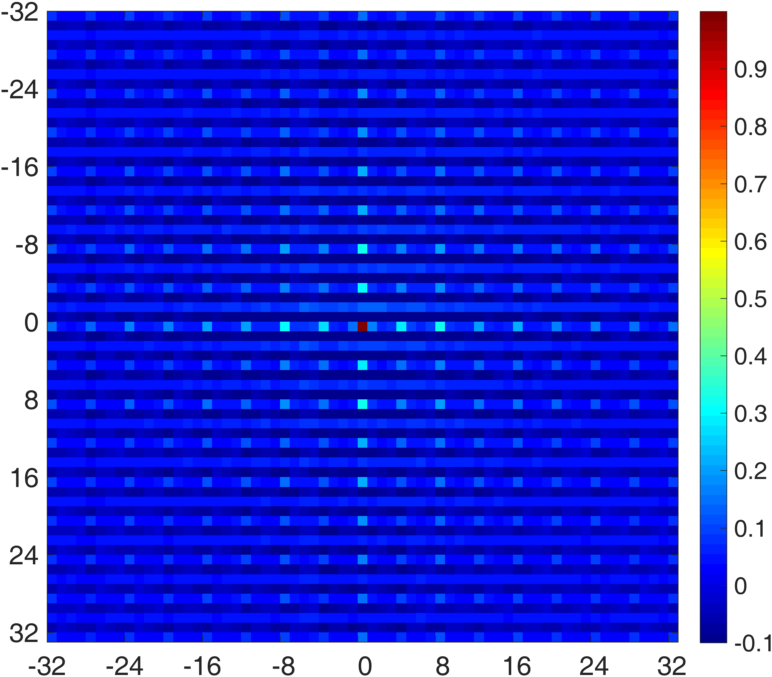} &
		\includegraphics[width=40mm,height=36mm]{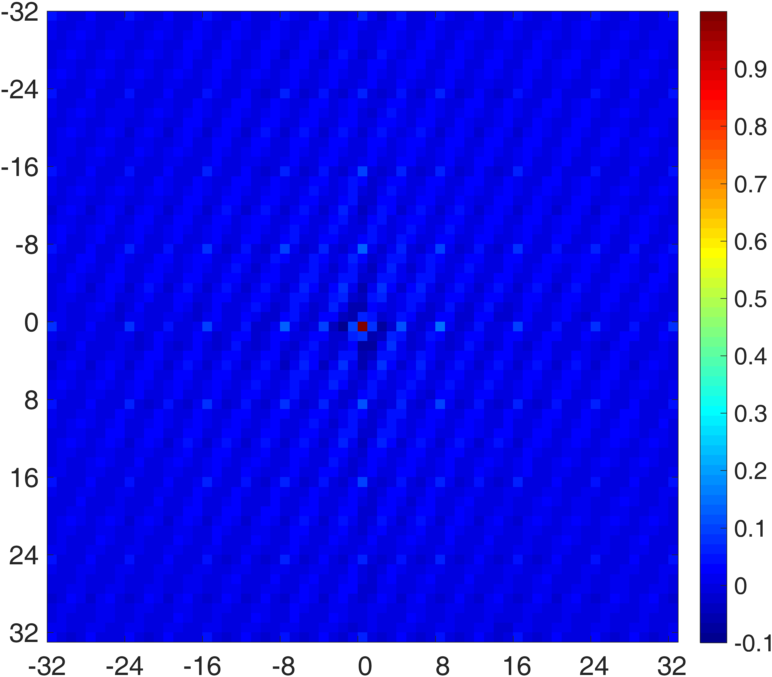}
    \end{tabular}
	\caption{\small Autocorrelation matrices of the Cycle-GAN and Pro-GAN fingerprints ($N$=512) of Figure 2.}
	\label{fig:Autocorrelations}	
\end{figure}

We now take a more application-oriented point of view,
looking for these fingerprints' ability to tell apart images of different origin.
Based on image-to-fingerprint correlation or similar indicators,
meaningful fingerprints should allow one to decide which of the two GANs generated a given image.

\newcommand{\tX}{\widetilde{X}}
\newcommand{\tY}{\widetilde{Y}}
Let
\begin{equation}
    {\rm corr}(X,Y) =  \tX \odot \tY
\end{equation}
be the correlation index between images $X$ and $Y$, where $\tX$ is the zero-mean unit-norm version of $X$ and $\odot$ indicates inner product.
For both GANs under analysis,
we regard the estimates obtained with $N=2^{9}$ as the ``true'' fingerprints, $F_A$ and $F_B$, respectively.
Then, we compute the correlation indices
between residuals $R^A_i, i=1,\ldots,M$ generated by GAN $A$ (and not used to estimate the fingerprint),
and the same-GAN ($F_A$) and cross-GAN ($F_B$) fingerprints,
that is
\begin{equation}
    \rho^A_{i,{\rm same}} = {\rm corr}(R^A_i,F_A)
\end{equation}
and
\begin{equation}
    \rho^A_{i,{\rm cross}} = {\rm corr}(R^A_i,F_B)
\end{equation}

\begin{figure}
	\centering
	\begin{tabular}{c@{\hspace{2mm}}c}
		\includegraphics[width=40mm,height=36mm]{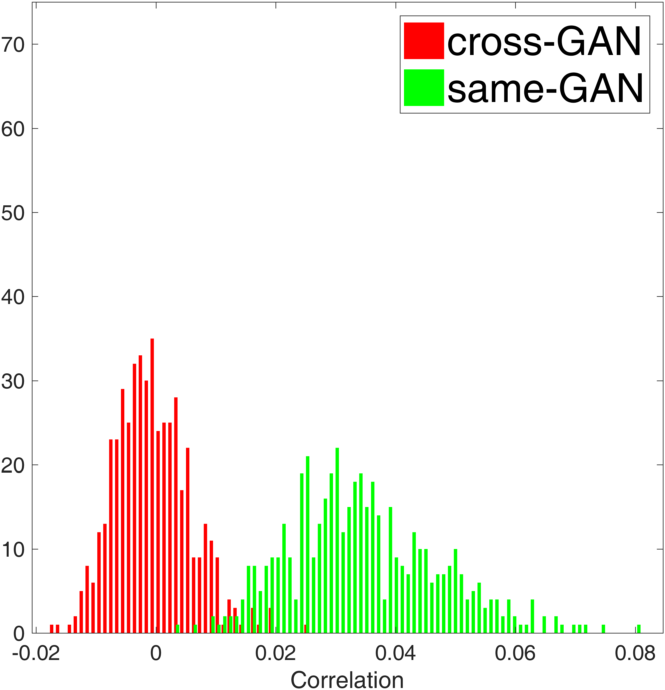} &
		\includegraphics[width=40mm,height=36mm]{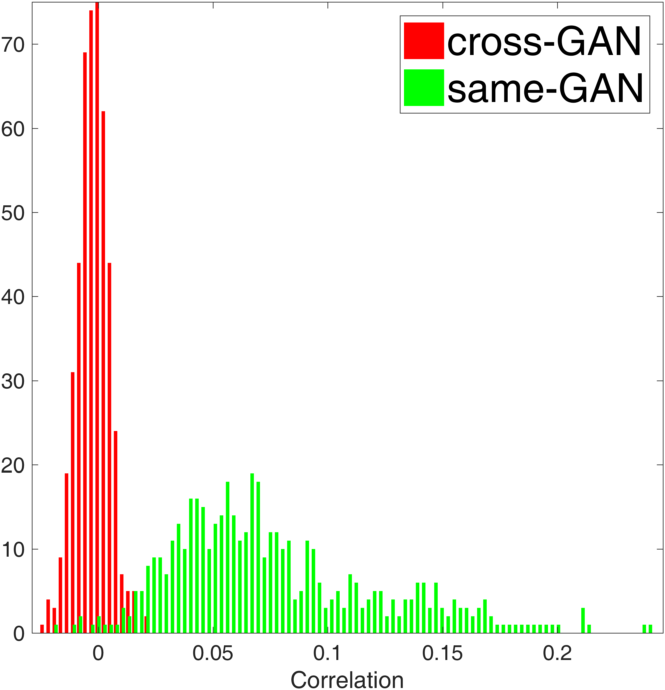}
    \end{tabular}
	\caption{\small Correlation of Cycle-GAN (left) and Pro-GAN (right) residuals with same/cross-GAN fingerprints.}
	\label{fig:Rho}	
\end{figure}

Fig.5(left) shows the histograms of same-GAN (green) and cross-GAN (red) correlations.
Cross-GAN correlations are evenly distributed around zero, indicating no correlation between generated images and the unrelated fingerprint.
On the contrary, same-GAN correlations are markedly larger than zero, testifying of a significant correlation with the correct fingerprint.
The behavior is very similar when GAN-B residuals are considered and the roles are reversed, see Fig.5(right).
Moreover, in both cases the two distributions are well separated, allowing reliable discrimination.
The corresponding receiver operating curves (ROC) are nearly perfect with area under curve (AUC) 0.990 and 0.998, respectively.

We carried out similar experiments for many other GANs, differing for architecture and/or training set, obtaining always similar results.
These results provide a convincing answer to our fundamental question,
showing that each GAN leaves a distinctive mark on each image generated by it, which can be legitimately called fingerprint.

\section{Source identification experiments}

Let us now consider a more challenging experimental setting, to carry out larger-scale source identification tests.
We consider three GAN architectures, Cycle-GAN, Pro-GAN, and Star-GAN.
Cycle-GAN was proposed in \cite{Zhu2017} to perform image-to-image translation.
The generator takes an input image of the source domain and transforms it into a new image of the target domain ({\it e.g.}, apples to oranges).
To improve the photorealism of generated images, a cycle consistency constraint is enforced.
Here, we consider several Cycle-GAN networks, trained by the authors on different source/target domains.
The second architecture, Progressive-GAN \cite{karras2018},
uses progressively growing generator and discriminator to create images of arbitrary-size which mimic images of the target domain.
In this case too, six different target domains are considered.
Like Cycle-GAN, Star-GAN \cite{Choi2018} performs image-to-image translation, but adopts a unified approach such that
a single generator is trained to map an input image to one of multiple target domains, which can be selected by the user.
By sharing the generator weights among different domains, a dramatic reduction of the number of parameters is achieved.
Finally, we include also two sets, from the RAISE dataset \cite{RAISE}, of images acquired by real cameras, so as to compare the behavior of real-world and GAN fingerprints.
Table I lists all networks and cameras, with corresponding abbreviations.
For each dataset ${\cal A}$, we generate/take 1000 RGB images of 256$\times$256 pixels,
extract residuals, and use $N$=512 of them to estimate the fingerprint $F_A$, and the remaining $M$=488, $\{R^A_1,\ldots,R^A_M\}$ for testing.

\begin{table}[h]
\centering
{\footnotesize
\begin{tabular}{|c|l|c|} \hline
Architecture \rule{0mm}{3mm} & Target / Camera model & Abbreviation \\ \hline
\multirow{9}{*}{Cycle-GAN}   & apple2orange          & C1           \\
                             & horse2zebra           & C2           \\
                             & monet2photo           & C3           \\
                             & orange2apple          & C4           \\
                             & photo2Cezanne         & C5           \\
                             & photo2Monet           & C6           \\
                             & photo2Ukiyoe          & C7           \\
                             & photo2VanGogh         & C8           \\
                             & zebra2horse           & C9           \\ \hline
\multirow{6}{*}{Pro-GAN}     & bedroom               & G1           \\
                             & bridge                & G2           \\
                             & church                & G3           \\
                             & kitchen               & G4           \\
                             & tower                 & G5           \\
                             & celebA                & G6           \\ \hline
\multirow{5}{*}{Star-GAN}    & black hair            & S1           \\
                             & blond hair            & S2           \\
                             & brown hair            & S3           \\
                             & male                  & S4           \\
                             & smiling               & S5           \\ \hline
\multirow{2}{*}{n/a}         & Nikon-D90             & N1           \\
                             & Nikon-D7000           & N2           \\ \hline
\end{tabular}
}
\caption{\small Cameras and GANs used in the experiments}
\label{tab:dataset}%
\end{table}%

\begin{figure}
	\centering
		\includegraphics[width=84mm,height=80mm]{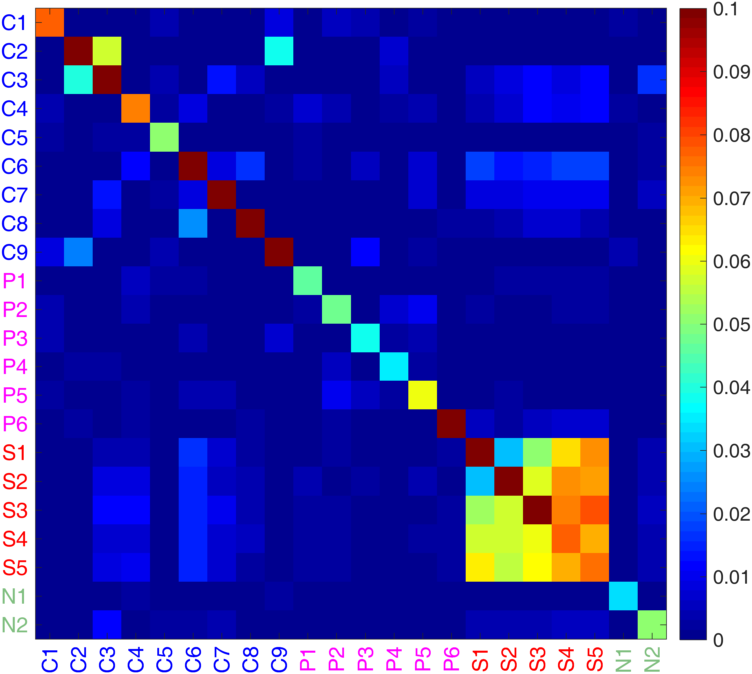}
	\caption{\small Average residual-fingerprint correlations.}
	\label{fig:correlations}	
\end{figure}

First of all we compute the average correlation between all sets of residuals and all fingerprints, that is
\begin{equation}
    \langle\rho\rangle^A_B = \frac{1}{M} \sum_{i=1}^M {\rm corr}(R^A_i,F_B)
\end{equation}
with $A,B$ spanning all sets.
Fig.6 shows a false-color representation of all such correlations.
It appears that diagonal entries are much larger, in general, than off-diagonal ones,
confirming that residuals of a dataset correlate well only with the fingerprint of the same dataset, be it GAN or natural.
There is also a clear block structure,
showing that some (weaker) correlation exists between residuals of a dataset
and fingerprints of ``sibling'' datasets, associated with the same GAN architecture.
This holds especially for the Star-GAN datasets, since the weights of a single generator are shared among all target domains.
Finally, as expected, no significant correlation exists between real and GAN-generated images,
which can be told apart easily based on their respective fingerprints.

We now perform camera attribution.
For each image, we compute the distance between the corresponding residual and all fingerprints, attributing the image with a minimum-distance rule.
In Fig.7 we show the resulting ROCs, and in Fig.8 the confusion matrix (entries below 1\% are canceled to improve readability).
Despite the 2$\times$ zooming, the ROC figure is hard to read as all curves amass in the upper-left corner.
On the other hand, this is the only real message we wanted to gather from this figure:
attribution is very accurate in all cases, with the only exception of the Star-GAN male and smiling networks.
This observation is reinforced by the confusion matrix, showing almost perfect attribution in all cases (with the same exceptions as before),
and with a slightly worse performance for the real cameras, characterized by a lower-energy PRNU.
Since real cameras usually compress images at high quality to save storage space,
we also repeated the attribution experiment after JPEG compressing all GAN-generated images at QF=95,
observing a negligible loss in accuracy, from 90.3\% to 90.1\%.

\begin{figure}
	\centering
	\begin{tabular}{cc}
		\includegraphics[width=75mm,height=75mm]{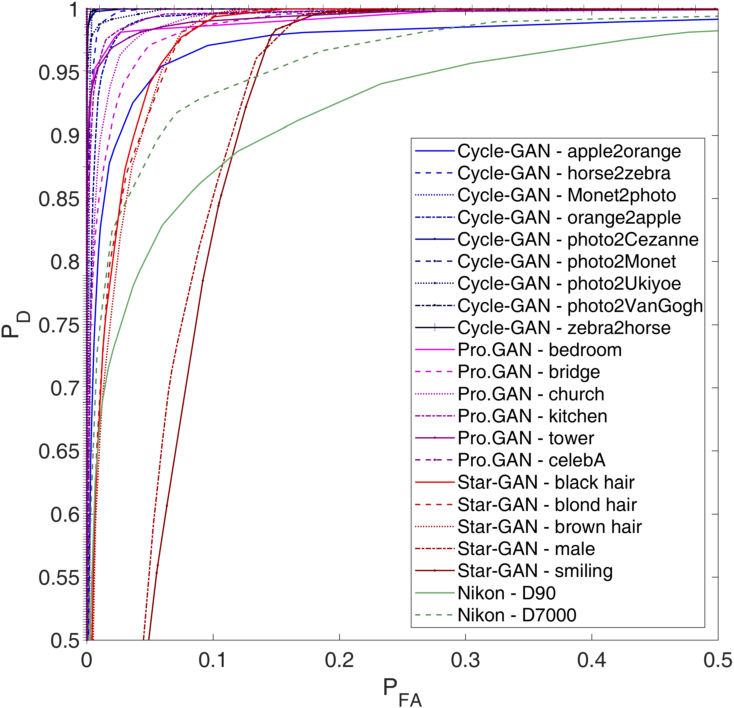}
    \end{tabular}
	\caption{\small One-vs.-all source identification ROCs.}
	\label{fig:ROCs}	
\end{figure}

\begin{figure*}
	\centering
	\begin{tabular}{cc}
		\includegraphics[width=160mm,height=80mm]{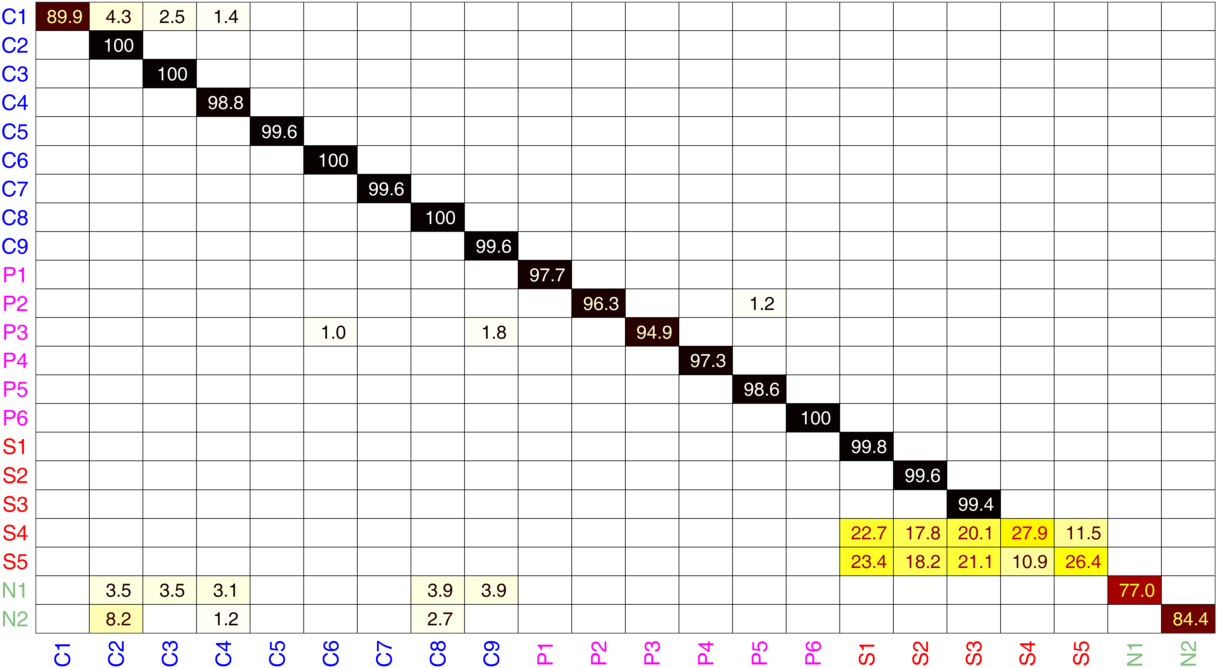}
    \end{tabular}
	\caption{\small Source identification confusion matrix. Entries below 1\% are canceled.}
	\label{fig:Confusion}	
\end{figure*}

We conclude this Section by reporting very briefly on the results obtained in the ``Forensics GAN Challenge'' \cite{GANchallenge2018}
organized in June-July 2018 by the US National Institute of Standards and Technology in the context of the Medifor program.
The goal was to classify as real or GAN-generated 1000 images of widely different resolution, from 52$\times$256 to 4608$\times$3072 pixels.
As baseline method we used a deep network trained on a large number of images retrieved from the InterNet.
However, we also tested the GAN fingerprint idea, following the scheme outlined in Fig.9.
We computed fingerprints for several popular GANs and, eventually,
identified a large cluster of size-1024$\times$1024 images generated with the same GAN.
This allowed us to improve the deep net accuracy by a simple fusion rule, for a final 0.999 AUC.

\begin{figure}[t!]
	\centering
\includegraphics[width=84mm,height=50mm]{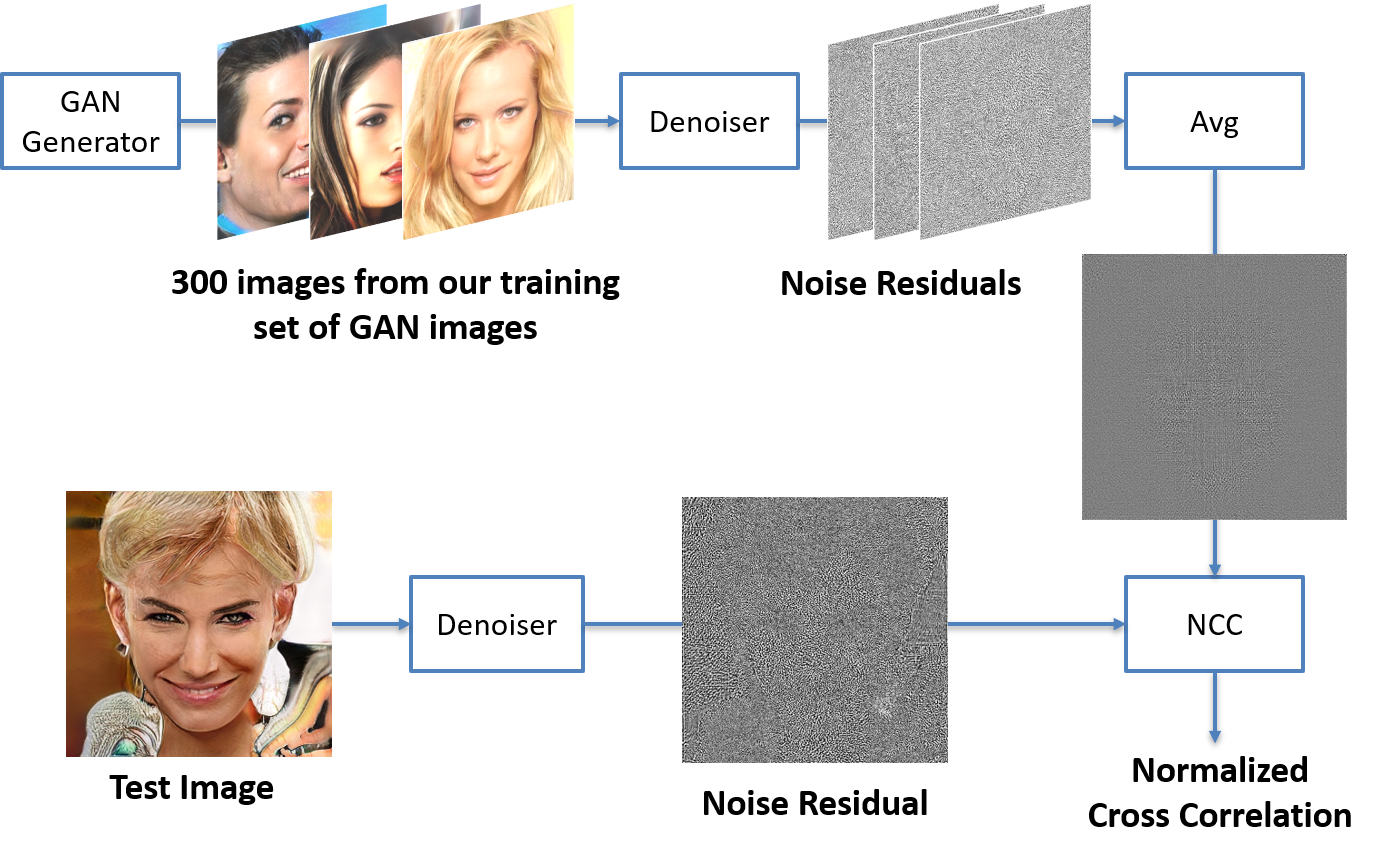}
	\caption{\small GAN-fingerprints for the Forensics Challenge.}
	\label{fig:Challenge}	
\end{figure}

\section{Conclusions and future work}
The goal of this work was to demonstrate the existence of GAN fingerprints and their value for reliable forensic analyses.
We believe both facts are supported by a sufficient experimental evidence.
This answers our fundamental question, but introduces many more questions and interesting topics which deserve further investigation.

First of all,
it is important to understand how the fingerprint depends on the network, both its architecture (number and type of layers) and its specific parameters (filter weights).
This may allow one to improve the fingerprint quality
or, with the attacker's point of view, find ways to remove the fingerprint from generated images as a counter-forensic measure.
Along the same path,
our preliminary results suggest that training the same architecture with different datasets gives rise to well distinct fingerprints.
Is this true in general? Will fine-tuning produce similar effects?

Under a more practical point of view,
further studies are necessary to assess the potential of GAN fingerprints in multimedia forensics. Major aims, besides source identification, are the discrimination between real and GAN-generated images, and the localization of GAN-generated material spliced in real images.
It is also important to study the robustness of such fingerprints to subsequent processing, such as JPEG compression, resizing, blurring, noising.
Finally, it is worth assessing the dependence of performance on the number and size of images used for fingerprint estimation,
with blind attribution and clustering as an interesting limiting case. 

\balance
\bibliographystyle{latex8}
\bibliography{refs}

\end{document}